\documentclass[10pt,journal,epsfig]{IEEEtran}

\usepackage{graphicx}
\usepackage{amssymb}
\usepackage{cite}
\usepackage{subfigure}
\usepackage{amsmath}
\usepackage{float}
\usepackage{booktabs}

\usepackage{algorithm}
\usepackage{algorithmic}

\begin{document}
\title{Low-Rank Phase Retrieval via Variational Bayesian Learning}

\author{Kaihui Liu, Jiayi Wang, Zhengli Xing, Linxiao Yang, and Jun Fang
     \thanks{Kaihui Liu, Jiayi
Wang, and Linxiao Yang, and Jun Fang 
are with the National Key Laboratory of Science and Technology on Communications, University
of Electronic Science and Technology of China, Chengdu 611731,
China (e-mail: kailiu@std.uestc.edu.cn).}
\thanks{Zhengli Xing is with China Academy of Engineering Physics, Mianyang
621900, China.}
}
\maketitle

\begin{abstract}
In this paper, we consider the problem of low-rank phase retrieval
whose objective is to estimate a complex low-rank matrix from
magnitude-only measurements. We propose a hierarchical prior model
for low-rank phase retrieval, in which a Gaussian-Wishart
    hierarchical prior is placed on the underlying low-rank matrix to
    promote the low-rankness of the matrix. Based on the proposed
    hierarchical model, a variational expectation-maximization (EM)
    algorithm is developed. The proposed method is less sensitive
    to the choice of the initialization point and works well with random initialization.
    Simulation results are provided to
    illustrate the effectiveness of the proposed algorithm.
\end{abstract}



\section{Introduction}
Phase retrieval has attracted a lot of interest for its practical
significance in many applications, including X-ray crystallography
\cite{Harrison93,Millane90}, optics \cite{Walther63}, astronomy
\cite{FienupDainty87}, diffraction imaging
\cite{Bunk2007Diffractive}, acoustics \cite{BalanCasazza06},
quantum mechanics \cite{Corbett2006The} and quantum information
\cite{Heinosaari2013Quantum}, etc. In these applications, the
phase information of measurements is difficult to obtain or
completely missing, and one hope to recover the unknown signal
from only the intensity of the measurements. The problem of
recovering a complex signal $\boldsymbol{x}\in \mathbb{C}^{n}$
from magnitude-only measurements $\boldsymbol{y}\in
\mathbb{R}^{m}$ can be cast as
\begin{equation}
\boldsymbol{y} = |\boldsymbol{\Phi}\boldsymbol{x}|
\label{model}
\end{equation}
where $\boldsymbol{\Phi}\in \mathbb{C}^{m \times n}$ is a known
complex measurement matrix.






A variety of phase retrieval algorithms have been proposed over
the past few years. One of the earliest works on phase retrieval
is the Gerchberg-Saxton algorithm \cite{Gerchberg72}, which is
based on alternating projection and iterates between the unknown
phases of the measurements and the unknown signals. In
\cite{NetrapalliJain13}, an alternating minimization technique
with spectral initialization was developed. Recently, a convex
relaxation-based method which enjoys a nice theoretical guarantee,
named PhaseLift, was proposed for phase retrieval in
\cite{CandesStrohmer13}. PhaseLift transforms phase retrieval into
a semidefinite program via convex relaxation. But, in the
meanwhile, the signal of interest is lifted to a high dimensional
space and thus the PhaseLift incurs a higher computational
complexity. To avoid this drawback, nonconvex phase retrieval
algorithms that directly work on the original signal space were
developed, e.g. alternating minimization algorithm
\cite{NetrapalliJain13}, Wirtinger flow algorithm
\cite{CandesLi15} and its variants
\cite{ChenCandes17,YuanWang17,zhang2017nonconvex,ma2017implicit,
Yang2017Misspecified,WangGiannakis18,wang2018phase,pinilla2018phase},
Almost all of these nonconvex methods require a sufficiently
accurate initial point to guarantee to find a globally optimal
solution. One exception is the gradient descent method with random
initialization proposed in \cite{ChenChi18}, which works as long
as the sample complexity exceeds $m\gtrsim n\textrm{poly}\log(n)$.
More recently, a non-lifting convex relaxation-based method,
called PhaseMax
\cite{bahmani2017flexible,GoldsteinStuder18,Hand2016An} was
proposed for phase retrieval, which relaxed the nonconvex equality
constraints in (\ref{model}) to convex inequality constraints, and
solved a linear program in the natural parameter space.




In recent years, signals with more complex structures have been
considered in phase retrieval problems. For example, in many
applications, the signal of interest is sparse or approximately
sparse, and this sparse structure can be utilized to significantly
reduce the sample complexity for exact phase recovery
\cite{MoravecRomberg07,ShechtmanBeck14,LiVoroninski13,SchniterRangan15,
eldar2015sparse,cai2016optimal,iwen2017robust,pedarsani2017phasecode,JagatapHegde17,Wang2018Sparse,hand2018phase}.
In some other applications, one may collect intensity measurements
of a time-varying signal at different time instants, thus
resulting in a multiple measurement vector (MMV) model. Due to the
temporal correlation between different time instants, the signal
of interest in a matrix form usually exhibits a low-rank structure
\cite{VaswaniNayer17}. Such a low-rank phase retrieval problem was
firstly introduced in \cite{VaswaniNayer17}, where an alternating
minimization method was developed. In this paper, we address the
low-rank phase retrieval problem from a Bayesian perspective. In
particular, we propose a hierarchical prior model for low-rank
phase retrieval, in which a Gaussian-Wishart hierarchical prior
originally introduced in \cite{YangFang18} is employed to promote
the low-rankness of the matrix. Based on the hierarchical Bayesian
model, a variational expectation-maximization (EM) algorithm is
developed with phase information treated as unknown deterministic
parameters. The proposed method is The proposed method is less
sensitive to the choice of the initialization point and works well
with random initialization. Simulation results show that our
method is capable of obtaining efficient and robust results.

The rest of the paper is organized as follows. In Section II, a
hierarchical Gaussian prior model for low-rank phase retrieval is
introduced. Based on this hierarchical model, a variational
Bayesian learning method is developed in Section III. Simulation
results are provided in Section IV,  followed by concluding
remarks in Section V.





\section{Bayesian Modeling}
We consider a low-rank phase retrieval problem which was firstly
introduced in \cite{VaswaniNayer17}. Suppose we have a low-rank
matrix $\boldsymbol{X}\triangleq
[\boldsymbol{x}_1\phantom{0}\ldots\phantom{0}\boldsymbol{x}_M]\in\mathbb{C}^{N\times
    M}$ with its rank $r \ll \min\{N,M\}$. For each column
$\boldsymbol{x}_m$ of the low-rank matrix, we can collect $P$
non-linear intensity measurements of the form
\begin{align}
y_{p,m}=|\boldsymbol{a}_{p,m}^H \boldsymbol{x}_m+w_{p,m}|,\quad
p=1,\ldots, P,\;\; m=1,\ldots, M
\label{lrm}
\end{align}
where $\boldsymbol{a}_{p,m}\in\mathbb{C}^{N}$ is the measurement
vector, and $w_{p,m}$ denotes the additive noise. The problem of
interest is to recover the low-rank matrix $\boldsymbol{X}$ from
the $P\times M$ phaseless measurements $\{y_{p,m}\}$. Since we
only have the magnitude measurements, each column of the low-rank
matrix, $\boldsymbol{x}_m$, can only be recovered up to a phase
ambiguity. Naturally, this low-rank phase retrieval problem can be
formulated as
\begin{align}
\min_{\boldsymbol{X}}\quad& \text{rank}(\boldsymbol{X})\nonumber \\
\text{s.t.}\quad& \boldsymbol{y}_m=|\boldsymbol{A}_m
\boldsymbol{x}_m+\boldsymbol{w}_m| \quad \forall m \label{opt1}
\end{align}
where $\boldsymbol{y}_m\triangleq
[y_{1,m}\phantom{0}\ldots\phantom{0}y_{P,m}]^T$,
$\boldsymbol{w}_m\triangleq
[w_{1,m}\phantom{0}\ldots\phantom{0}w_{P,m}]^T$, and
$\boldsymbol{A}_m\triangleq
[\boldsymbol{a}_{1,m}\phantom{0}\ldots\phantom{0}\boldsymbol{a}_{P,m}]^H$.
The optimization (\ref{opt1}), however, is difficult to solve due
to its non-convexity. In this paper, we consider modeling the
low-rank phase retrieval problem within a Bayesian framework. To
facilitate our modeling, similar to \cite{DremeauKrzakala15}, we
introduce a set of deterministic parameters $\{\theta_{p,m}\}$,
where $\theta_{p,m}\in [0,2\pi)$ represents the missing phase of
the observation $\tilde{y}_{p,m}\triangleq\boldsymbol{a}_{p,m}^H
\boldsymbol{x}_m +w_{p,m}$. Therefore we can write
\begin{align}
y_{p,m}=e^{-j\theta_{p,m}}(\boldsymbol{a}_{p,m}^H \boldsymbol{x}_m
+w_{p,m}) \quad \forall p, \forall m
\end{align}
and the measurements in a vector form can be written as
\begin{align}
\boldsymbol{y}_m=\boldsymbol{D}_m(\boldsymbol{A}_m
\boldsymbol{x}_m+\boldsymbol{w}_m) \quad  \forall m
\end{align}
where
\begin{align}
\boldsymbol{D}_m\triangleq\text{diag}(e^{-j\theta_{1,m}},\ldots,e^{-j\theta_{P,m}})
\end{align}
We assume entries of $\boldsymbol{w}_m$ are independent and
identically distributed (i.i.d.) random variables following a
Gaussian distribution with zero mean and variance $\beta^{-1}$. To
learn $\beta$, we place a Gamma hyperprior over $\beta$, i.e.
\begin{align}
p(\beta) = \text{Gamma}(\beta|a,b) = \Gamma(a)^{-1} b^a
\beta^{a-1} e^{-b\beta}
\end{align}
where $\Gamma(a) = \int_{0}^{\infty} t^{a-1}e^{-t}dt$ is the Gamma
function. The parameters $a$ and $b$ are set to be small values,
e.g. $10^{-10}$, which makes the Gamma distribution a
non-informative prior.

To promote a low-rank solution of $\boldsymbol{X}$, we place a
two-layer hierarchical Gaussian prior \cite{YangFang18} on
$\boldsymbol{X}$. It was shown in \cite{YangFang18} this two-layer
hierarchical Gaussian prior model has the potential to encourage a
low-rank solution. Specifically, in the first layer, the columns
of $\boldsymbol{X}$ are assumed mutually independent and follow a
common Gaussian distribution:
\begin{align}
p(\boldsymbol{X}|\boldsymbol{\Sigma})=\prod\limits_{m=1}^{M}p(\boldsymbol{x}_m|\boldsymbol{\Sigma})
=\prod\limits_{m=1}^{M}\mathcal{N}(\boldsymbol{x}_m|\boldsymbol{0},\boldsymbol{\Sigma}^{-1})\label{x-prior}
\end{align}
where $\boldsymbol{\Sigma}\in\mathbb{C}^{N\times N}$ is the
precision matrix. The second layer specifies a Wishart
distribution as a hyperprior over the precision matrix
$\boldsymbol{\Sigma}$:
\begin{align}
p(\boldsymbol{\Sigma}) \propto&
|\boldsymbol{\Sigma}|^{\frac{\nu-M-1}{2}}\exp(-\frac{1}{2}
\text{tr}(\boldsymbol{W}^{-1}\boldsymbol{\Sigma}))
\end{align}
where $\nu$ and $\boldsymbol{W}\in\mathbb{R}^{N\times N}$ denote
the degrees of freedom and the scale matrix of the Wishart
distribution, respectively. For the standard Wishart distribution,
$\nu$ is an integer which satisfies $\nu>N-1$. Nevertheless, if an
improper prior\footnote{In Bayesian inference, improper prior
    distributes can often be used provided that the corresponding
    posterior distribution can be correctly normalized
    \cite{Bishop07}.} is allowed \cite{PhilipMorris00}, the choice of
$\nu$ can be more flexible and relaxed as $\nu>0$.

\begin{figure}[t]
    \centering
    \includegraphics [width=220pt]{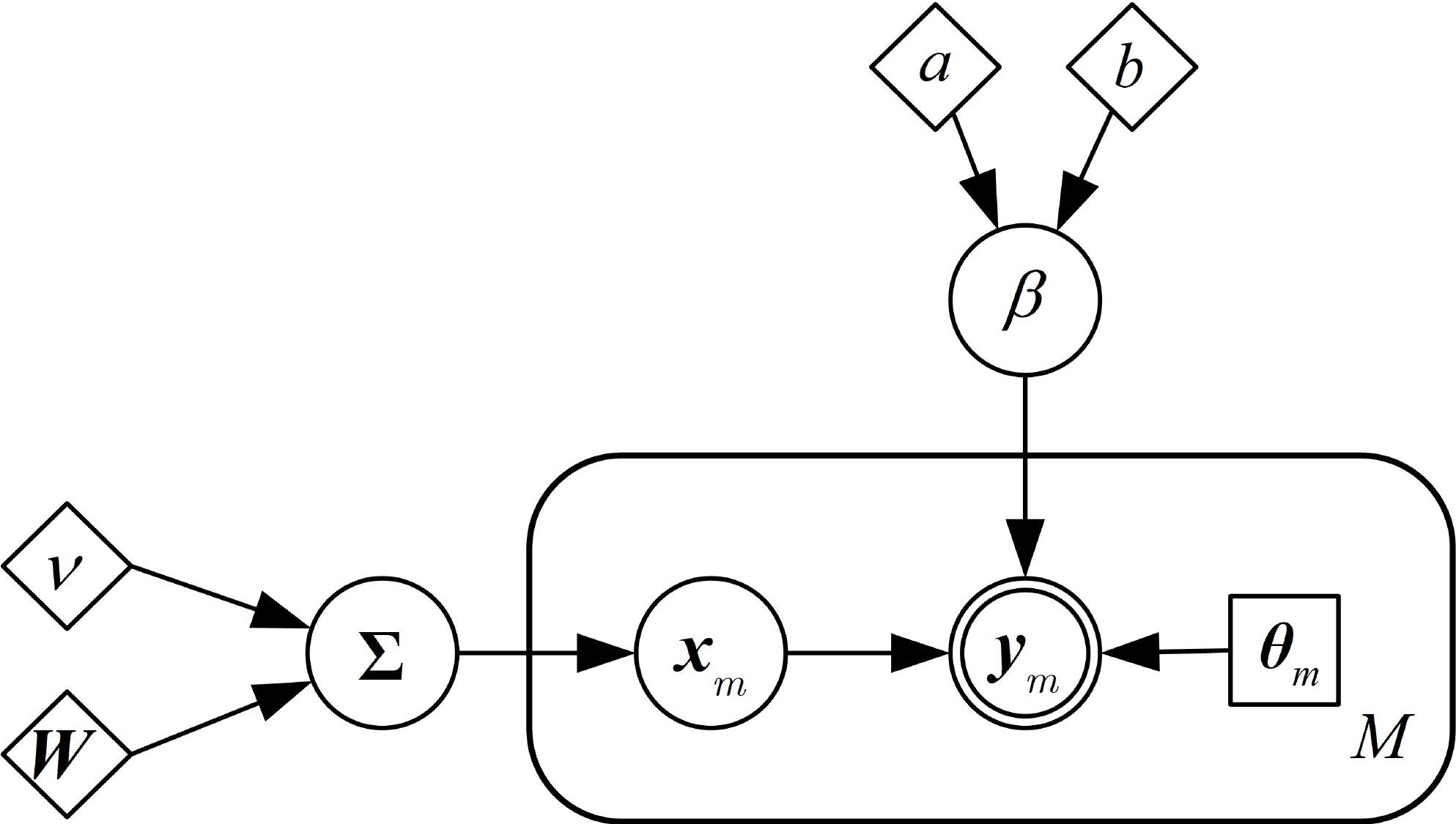}
    \caption{Graphical model for proposed Bayesian low-rank phase-retrieval
        method, in which
        double circles denote the observable variable, single
        circles denote the hidden variable,
        and the boxes denote deterministic hyperparameters,
        and the diamonds denote pre-specified hyperparameters}
    \label{fig:Graphical-Model}
\end{figure}

For clarity, we plot the graphical model for our low-rank
phase-retrieval problem in Fig. \ref{fig:Graphical-Model}.

\section{Variational Bayesian Inference}
\subsection{Review of The Variational Expectation Maximization
    Methodology} We first provide a brief review of the variational
expectation maximization methodology (VEM). More details about the
VEM can be found in \cite{TzikasLikas08}. In a probabilistic
model, let $\boldsymbol{t}$, $\boldsymbol{z}$ and
$\boldsymbol{\theta}$ denote the observed data, the hidden
variables and the deterministic parameters, respectively. It is
well-known that we can decompose the log-likelihood function into
sum of two following terms \cite{TzikasLikas08}
\begin{align}
\ln
p(\boldsymbol{t;\boldsymbol{\theta}})=L(q;\boldsymbol{\theta})+\text{KL}(q||
p) \label{variational-decomposition}
\end{align}
where
\begin{align}
L(q;\boldsymbol{\theta})=\int q(\boldsymbol{z})\ln
\frac{p(\boldsymbol{t},\boldsymbol{z};\boldsymbol{\theta})}{q(\boldsymbol{z})}d\boldsymbol{z}
\end{align}
and
\begin{align}
\text{KL}(q|| p)=-\int q(\boldsymbol{z})\ln
\frac{p(\boldsymbol{z}|\boldsymbol{t};\boldsymbol{\theta})}{q(\boldsymbol{z})}d\boldsymbol{z}
\end{align}
where $q(\boldsymbol{z})$ is an arbitrary probability density
function, $\text{KL}(q|| p)$ is the Kullback-Leibler divergence
between $p(\boldsymbol{z}|\boldsymbol{t};\boldsymbol{\theta})$ and
$q(\boldsymbol{z})$. Since $\text{KL}(q|| p)\geq 0$,
$L(q;\boldsymbol{\theta})$ is a lower bound on $\ln
p(\boldsymbol{t};\boldsymbol{\theta})$. It is usually difficult to
directly maximize the log-likelihood function $\ln
p(\boldsymbol{t};\boldsymbol{\theta})$. To address this
difficulty, we, alternatively, search for $\boldsymbol{\theta}$ by
maximizing its lower bound $L(q;\boldsymbol{\theta})$. This
naturally leads to a variational EM methodology, which
alternatively maximizes $L(q;\boldsymbol{\theta})$ with respect to
$q(\boldsymbol{z})$ and $\boldsymbol{\theta}$. First, given the
current estimate of $q^{\text{old}}(\boldsymbol{z})$, we update
$\boldsymbol{\theta}$ via maximizing
$L(q^{\text{old}};\boldsymbol{\theta})$, which is referred to as
the maximization step. Then, given the current estimate of
$\boldsymbol{\theta}^{\text{old}}$, we maximize
$L(q;\boldsymbol{\theta}^{\text{old}})$ with respect to
$q(\boldsymbol{z})$, which is referred to as the expectation step.
Specifically, to facilitate the optimization of
$L(q;\boldsymbol{\theta}^{\text{old}})$, a factorized form of
$q(\boldsymbol{z})$ is assumed \cite{TzikasLikas08}, i.e.
$q(\boldsymbol{z})=\prod_i q_i(z_i)$. The maximization of
$L(q;\boldsymbol{\theta}^{\text{old}})$ with respect to
$q(\boldsymbol{z})$ can be conducted in an alternating fashion for
each latent variable, which yields \cite{TzikasLikas08}
\begin{align}
q_i(z_i)=\frac{\exp(\langle\ln
    p(\boldsymbol{t},\boldsymbol{z};\boldsymbol{\theta}^{\text{old}})\rangle_{k\neq
        i})}{\int\exp(\langle\ln
    p(\boldsymbol{t},\boldsymbol{z};\boldsymbol{\theta}^{\text{old}})\rangle_{k\neq i})d\theta_i}
\label{general-update}
\end{align}
where $\langle\cdot\rangle_{k\neq i}$ denotes an expectation with
respect to the distributions $q_i(\theta_i)$ for all $k\neq i$.

\subsection{Proposed Algorithm}
We now proceed to perform Bayesian inference for our problem. Let
$\boldsymbol{z}\triangleq\{\boldsymbol{X},\boldsymbol{\Sigma},\beta\}$
denote the hidden variables, and
$\boldsymbol{\theta}\triangleq\{\theta_{p,m}\}$ denote the unknown
deterministic parameters in our graphical model. Our objective is
to obtain an estimate of $\boldsymbol{\theta}$, along with the
approximate posterior distribution for $\boldsymbol{z}$. We assume
that $q(\boldsymbol{z})$ has a factorized form over the hidden
variables $\{\boldsymbol{X},\boldsymbol{\Sigma},\beta\}$, i.e.
\begin{equation}
q(\boldsymbol{z})=q_X(\boldsymbol{X})q_{\Sigma}(\boldsymbol{\Sigma})q_\beta(\beta)
\end{equation}
As discussed in the previous subsection, the maximization of the
lower bound $L(q;\boldsymbol{\theta})$ can be conducted in an
alternating fashion, which leads to

\textbf{E-step:}
\begin{align*}
&\ln{q_X(\boldsymbol{X})} = \langle \ln{p(\boldsymbol{Y}|\boldsymbol{X},\beta)
    p(\boldsymbol{X}|\boldsymbol{\Sigma})}\rangle_{q_{\Sigma}(\boldsymbol{\Sigma})q_\beta(\beta)} + const,  \\
&\ln{q_\Sigma (\boldsymbol{\Sigma})} = \langle
\ln{p(\boldsymbol{X}|\boldsymbol{\Sigma})p(\boldsymbol{\Sigma})}\rangle_{q_X(\boldsymbol{X})q_{\beta}(\beta)} + const, \\
&\ln{q_\beta(\beta)} = \langle
\ln{p(\boldsymbol{Y}|\boldsymbol{X},\beta)p(\beta)} \rangle
_{q_{\Sigma}(\boldsymbol{\Sigma})q_{X}(\boldsymbol{X})} + const,
\end{align*}

\textbf{M-step:}
\begin{equation*}
\boldsymbol{\theta} = \arg \max_{\boldsymbol{\theta}} \langle
\ln{p(\boldsymbol{Y},\boldsymbol{X},\boldsymbol{\Sigma},\beta ;
    \boldsymbol{\theta})}
\rangle_{q_{\Sigma}(\boldsymbol{\Sigma})q_{X}(\boldsymbol{X})q_\beta(\beta)}
\end{equation*}
where $\langle\rangle_{q_1(\cdot)...q_K(\cdot)}$ denotes the
expectation with respect to the distributions
$\{q_k(\cdot)\}_{k=1}^K$. Details of the Bayesian inference are
provided next.

\textbf{E-Step: Update of $q_X(\boldsymbol{X})$}: Since columns of
$\boldsymbol{X}$ are mutually independent, the approximate
posterior distribution of each column can be calculated
independently as
\begin{align}
&\ln{q_x(\boldsymbol{x}_m)} \notag \\
&\propto \langle \ln{[p(\boldsymbol{y}_m|\boldsymbol{x}_m)
    p(\boldsymbol{x}_m|\boldsymbol{\Sigma})]} \rangle_{q_\Sigma(\boldsymbol{\Sigma})q_{\beta}(\beta)} \notag \\
&\propto \langle -\beta(\boldsymbol{D}_m^{-1}\boldsymbol{y}_m -
\boldsymbol{A}_m \boldsymbol{x}_m)^H(\boldsymbol{D}_m^{-1}\boldsymbol{y}_m -
\boldsymbol{A}_m \boldsymbol{x}_m)
- \boldsymbol{x}_m^H\boldsymbol{\Sigma}\boldsymbol{x}_m \rangle \notag \\
&\propto -\boldsymbol{x}_m^H(\langle \beta \rangle
\boldsymbol{A}_m^H\boldsymbol{A}_m \notag\\
&\;\;\;\;+ \langle \boldsymbol{\Sigma}
\rangle)\boldsymbol{x}_m + \langle \beta
\rangle(\boldsymbol{x}_m^H\boldsymbol{A}_m^H\boldsymbol{D}_m^{-1}\boldsymbol{y}_m
+(\boldsymbol{D}_m^{-1}\boldsymbol{y}_m)^H\boldsymbol{A}_m\boldsymbol{x}_m)\label{equ-1}
\end{align}
From (\ref{equ-1}), it can be seen that $\boldsymbol{x}_m$
follows a Gaussian distribution
\begin{equation}
q_x(\boldsymbol{x}_m) =
\mathcal{CN}(\boldsymbol{x}_m|\boldsymbol{\mu}_m,\boldsymbol{Q}_m)
\label{x-update}
\end{equation}
with its mean $\boldsymbol{\mu}_m$ and covariance matrix
$\boldsymbol{Q}_m$ given as
\begin{align}
\boldsymbol{\mu}_m &= \langle \beta \rangle \boldsymbol{Q}_m\boldsymbol{A}_m^H\boldsymbol{D}_m^{-1}\boldsymbol{y}_m \\
\boldsymbol{Q}_m &= (\langle\beta\rangle
\boldsymbol{A}_m^H\boldsymbol{A}_m + \langle \boldsymbol{\Sigma}
\rangle)^{-1}\label{eq_cov}
\end{align}

\textbf{E-Step: Update of $q_\Sigma(\boldsymbol{\Sigma})$}: The
approximate posterior $q_\Sigma(\boldsymbol{\Sigma})$ can be
obtained as
\begin{align}
&\ln {q_\Sigma(\boldsymbol{\Sigma})} \notag \\
&\propto \langle \ln{[\prod\limits_{m = 1}^{M} p(\boldsymbol{x}_m|\boldsymbol{\Sigma})
    p(\boldsymbol{\Sigma})]} \rangle_{q_X(\boldsymbol{X})} \notag \\
&\propto (\nu+M-N-1) \ln{|\boldsymbol{\Sigma}|} -
\text{tr}[(\boldsymbol{W}^{-1} + \langle \boldsymbol{XX}^H
\rangle)\boldsymbol{\Sigma}]
\end{align}
From the above equation, we see that the posterior of
$\boldsymbol{\Sigma}$ follows a Wishart distribution, that is
\begin{equation}
q_\Sigma(\boldsymbol{\Sigma}) =
\text{Wishart}(\boldsymbol{\Sigma};\hat{\boldsymbol{W}},\hat{\nu})
\label{Sigma-update}
\end{equation}
where
\begin{align}
\hat{\boldsymbol{W}} &= (\boldsymbol{W}^{-1} + \langle \boldsymbol{XX}^H \rangle)^{-1} \label{eq_wis}\\
\hat{\nu} &= \nu + M
\end{align}

\textbf{E-Step: Update of $q_\beta(\beta)$}: The variational
optimization of $q_\beta(\beta)$ yields
\begin{align}
&\ln {q_\beta(\beta)} \notag \\
&\propto \langle \ln{[\prod\limits_{m = 1}^{M} p(\boldsymbol{y}_m|\boldsymbol{x}_m,\beta)p(\beta)]}
\rangle _{q_{X}(\boldsymbol{X})} \notag \\
&\propto (PM + a - 1)\ln{\beta} \notag \\
&\;\;\;\;- (\sum\limits_{m=1}^{M}\langle
\|\boldsymbol{D}_m^{-1}\boldsymbol{y}_m - \boldsymbol{A}_m \boldsymbol{x}_m\|^2
\rangle + b ) \beta
\end{align}
It is easy to verify that $q_\beta(\beta)$ follows a Gamma
distribution
\begin{equation}
q_\beta(\beta) = \text{Gamma} (\beta|\hat{a},\hat{b})
\label{beta-update}
\end{equation}
where
\begin{align}
\hat{a} &= PM +a \\
\hat{b} &= \sum\limits_{m=1}^{M}\langle ||\boldsymbol{D}_m^{-1}\boldsymbol{y}_m
- \boldsymbol{A}_m \boldsymbol{x}_m||^2 \rangle + b
\end{align}

\textbf{M-Step: Update of $\boldsymbol{\theta}$}: We conduct the
partial derivative of $\langle
\ln{p(\boldsymbol{Y},\boldsymbol{X},\boldsymbol{\Sigma},\beta ;
    \boldsymbol{\theta})}
\rangle_{q_{\Sigma}(\boldsymbol{\Sigma})q_{X}(\boldsymbol{X})q_\beta(\beta)}$
with respect to $\theta_{p,m}$, which yields
\begin{align}
&\frac{\partial \langle \ln p(y_{p,m}|\boldsymbol{x}_m,\beta)
    \rangle_{q_X(\boldsymbol{X})q_\beta(\beta)}}{\partial \theta_{p,m}} \notag \\
&= -\langle \beta \rangle \langle \frac{\partial|y_{p,m}e^{j\theta_{p,m}} -
    \boldsymbol{a}_{p,m}^H\boldsymbol{x}_m|^2}{\partial \theta_{p,m}} \rangle \notag \\
&= -\langle \beta \rangle \langle \frac{\partial(y_{p,m}e^{j\theta_{p,m}} -
    \boldsymbol{a}_{p,m}^H\boldsymbol{x}_m)(y_{p,m}e^{-j\theta_{p,m}} -
    \boldsymbol{x}_m^H\boldsymbol{a}_{p,m})}{\partial \theta_{p,m}} \rangle \notag \\
&=-j\langle\beta\rangle y_{p,m}e^{-j\theta_{p,m}}
\langle \boldsymbol{a}_{p,m}^H\boldsymbol{x}_m - e^{2j\theta_{p,m}}\boldsymbol{x}_m^H\boldsymbol{a}_{p,m} \rangle
\nonumber\\
&= -j\langle\beta\rangle y_{p,m}e^{-j\theta_{p,m}}
(\boldsymbol{a}_{p,m}^H\boldsymbol{\mu}_m -
e^{2j\theta_{p,m}}\boldsymbol{\mu}_m^H\boldsymbol{a}_{p,m})
\end{align}

Setting the above partial derivative equal to zero, we have
\begin{align}
e^{2j\theta_{p,m}} =
\frac{\boldsymbol{a}_{p,m}^H\boldsymbol{\mu}_m}{\boldsymbol{\mu}_m^H\boldsymbol{a}_{p,m}}
\label{theta-update}
\end{align}
Then an estimate of $\theta_{p,m}$ can be obtained
\begin{align}
\theta_{p,m}=\frac{1}{2}\arctan\Big\{ \frac{\textrm{Im}
(\boldsymbol{\varUpsilon}_{p,m})}{\textrm{Re}(\boldsymbol{\varUpsilon}_{p,m})}\Big\}
\label{theta}
\end{align}
where $\boldsymbol{\varUpsilon}_{p,m}\triangleq \frac{\boldsymbol{a}_{p,m}^H
\boldsymbol{\mu}_m}{\boldsymbol{\mu}_m^H\boldsymbol{a}_{p,m}}$, $\textrm{Im}(.)$
and $\textrm{Re}(.)$ denote the imaginary and real parts of $(.)$, respectively.

For clarity, we summarize our algorithm as follows.

\begin{tabular*}{8.5cm}{l}
    \toprule
    \textbf{Algorithm 1}: LRPR-Variational EM \\
    \midrule
    \textbf{while} not converge \textbf{do} \\
    \textbf{E-step}: \\
    \quad Update $q_x(\boldsymbol{X})$ via (\ref{x-update})
    with $q_{\Sigma}(\boldsymbol{\Sigma})$ and $q_{\beta}(\beta)$ fixed; \\
    \quad Update $q_{\Sigma}(\boldsymbol{\Sigma})$ via (\ref{Sigma-update})
    with $q_x(\boldsymbol{X})$ and $q_{\beta}(\beta)$ fixed; \\
    \quad Update $q_{\beta}(\beta)$ via (\ref{beta-update})
    with $q_x(\boldsymbol{X})$ and $q_{\Sigma}(\boldsymbol{\Sigma})$ fixed;  \\
    \textbf{M-step}: \\
    \quad Estimate $\boldsymbol{\theta}$ via (\ref{theta-update}). \\
    \textbf{end while} \\
    \bottomrule
\end{tabular*}

\begin{figure*}[!t]
    \centering
    \subfigure[$P=500$]{\includegraphics[width=5.5cm]{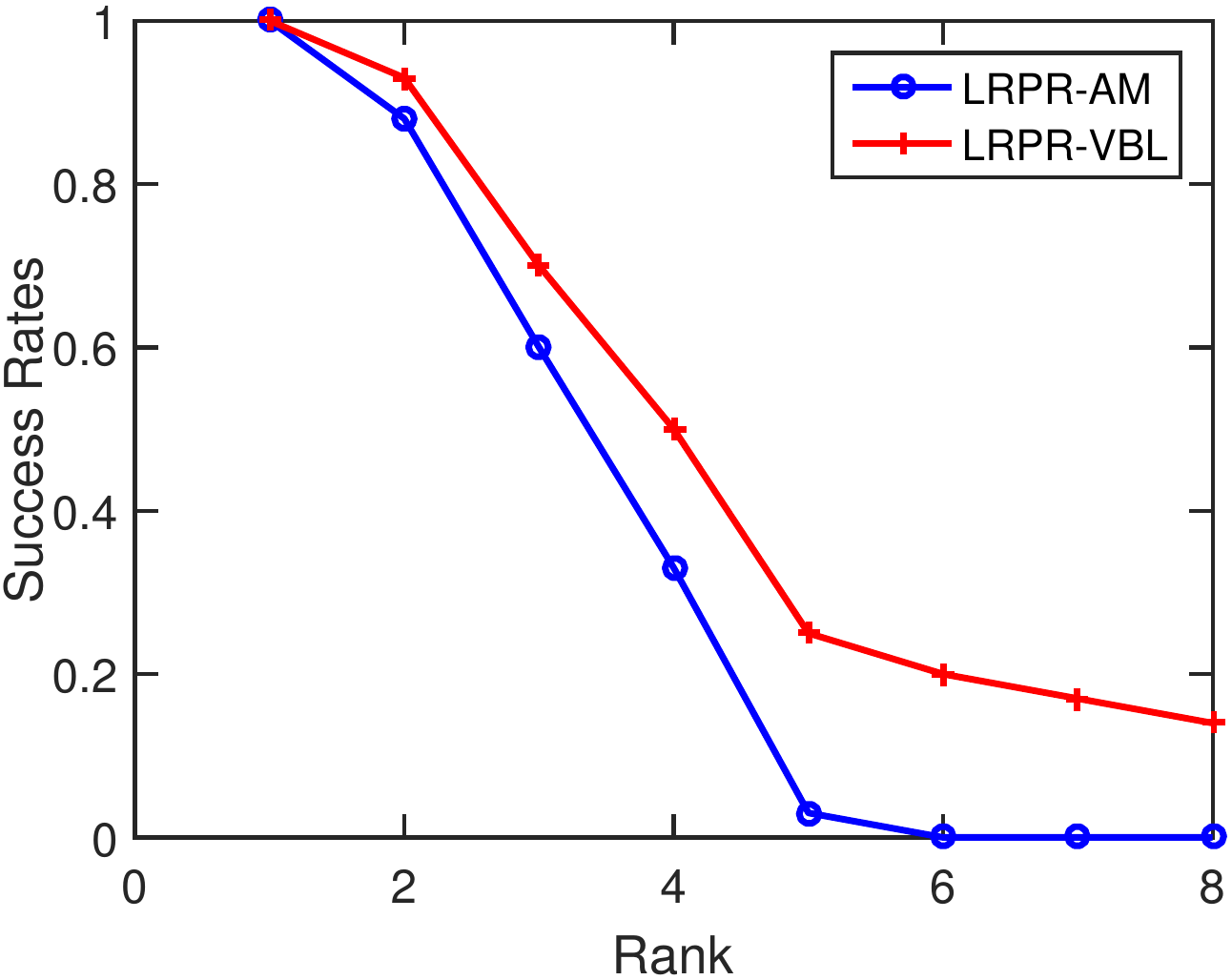}}
    \hfil
    \subfigure[$P=600$]{\includegraphics[width=5.5cm]{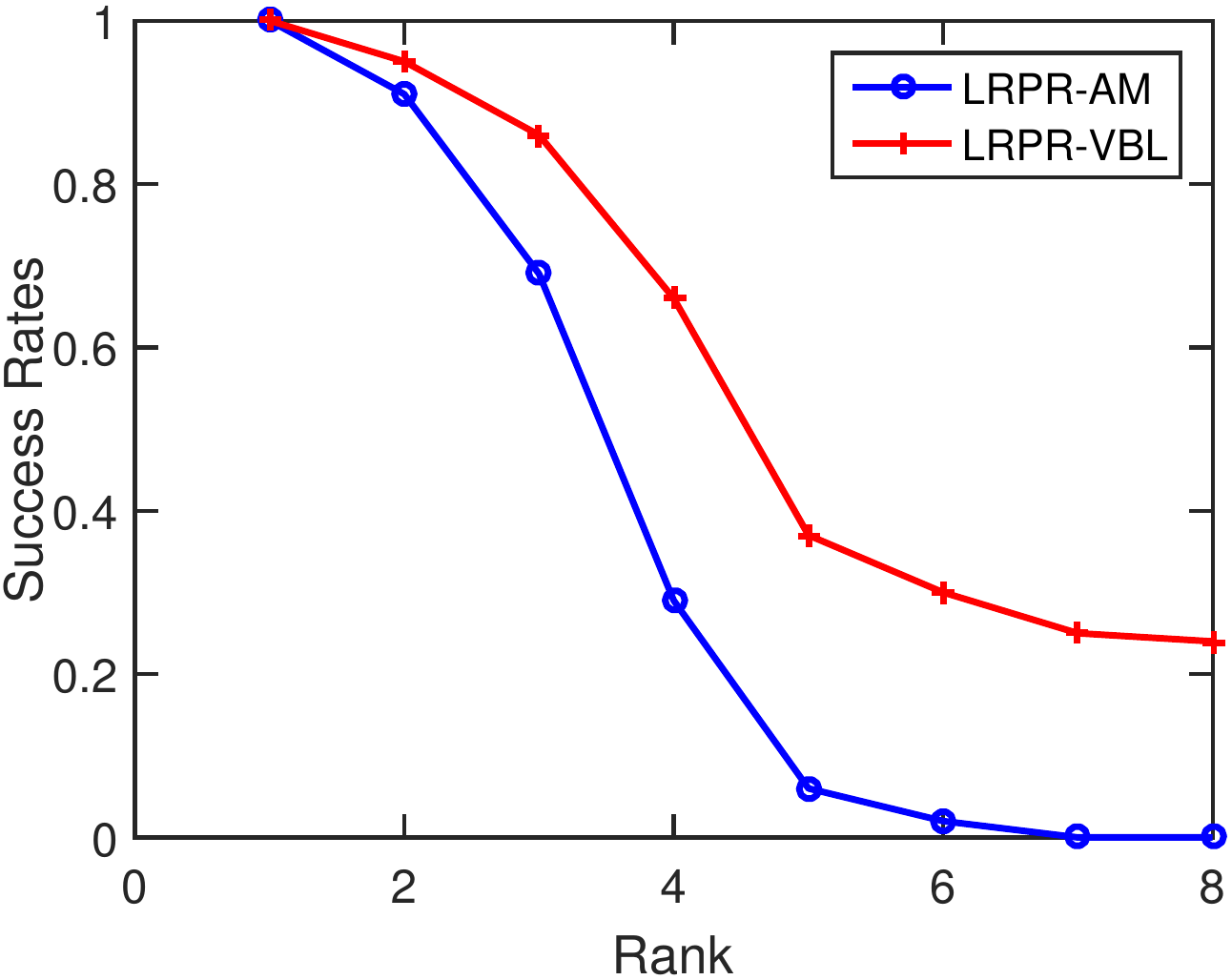}}
    \hfil
    \subfigure[$P=700$]{\includegraphics[width=5.5cm]{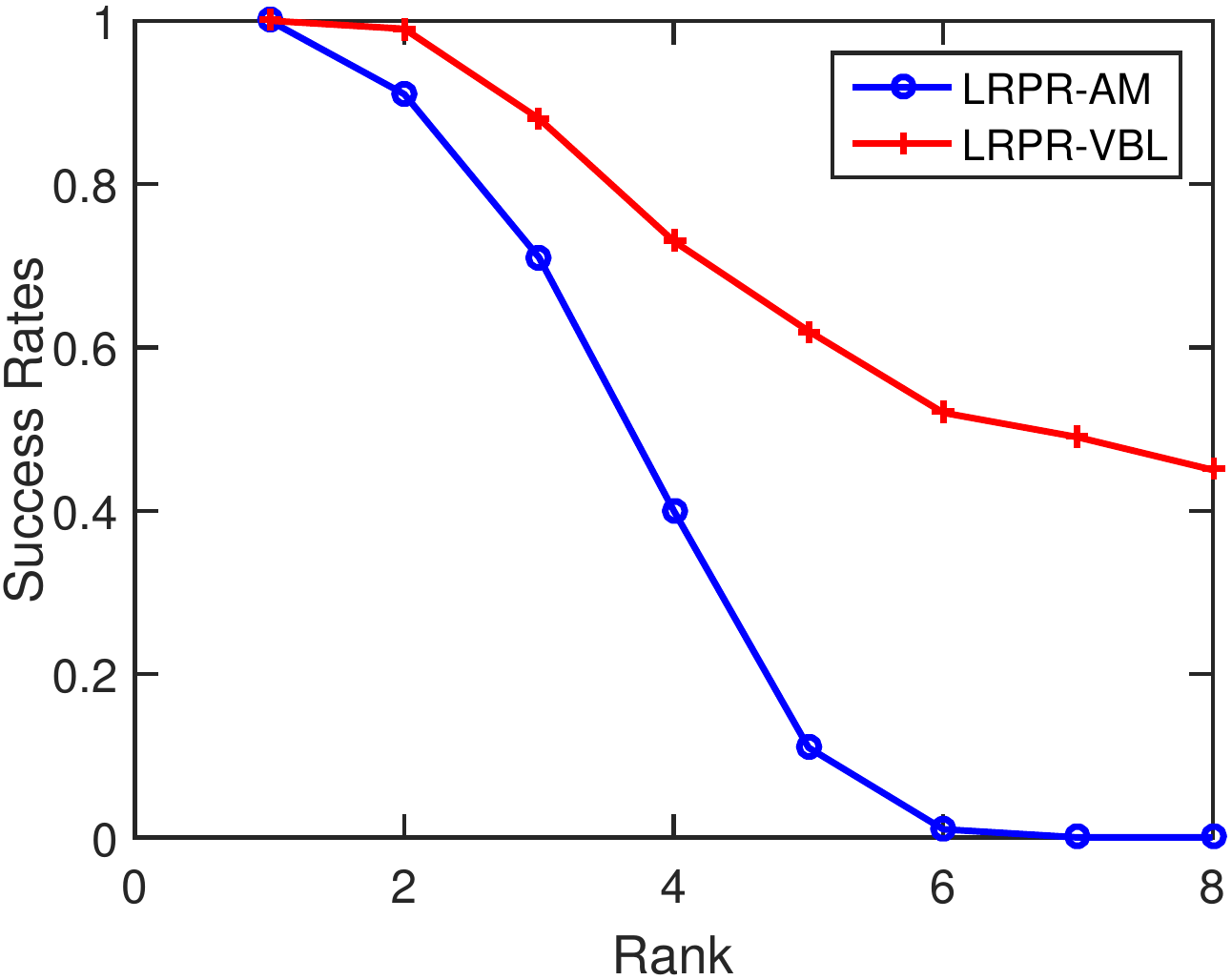}}
    \caption{Success rates of respective algorithms vs. the rank of the low-rank matrix $r$.}
    \label{fig1}
\end{figure*}

\begin{figure*}[!t]
    \centering
    \subfigure[$r=3$]{\includegraphics[width=5.5cm]{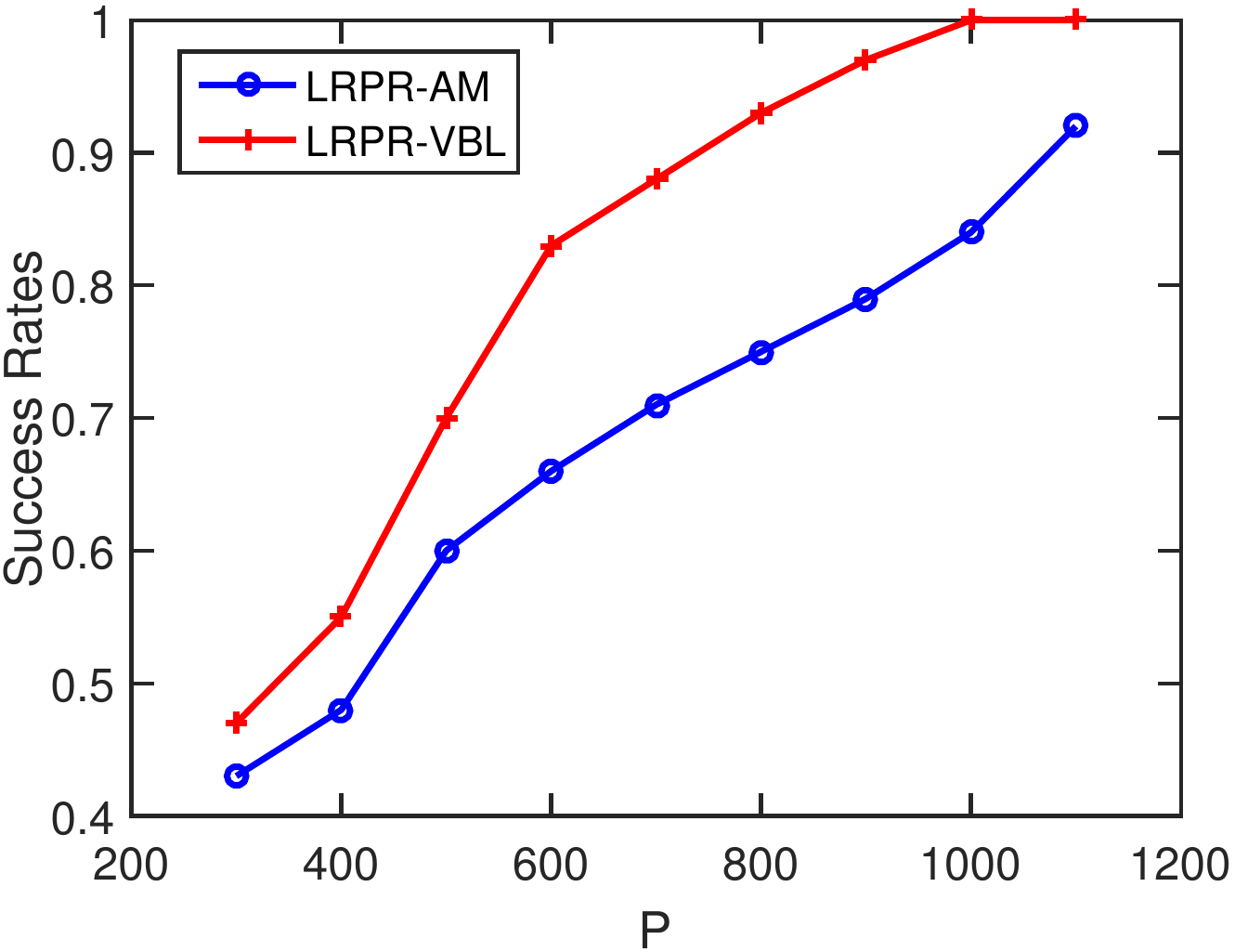}}
    \hfil
    \subfigure[$r=4$]{\includegraphics[width=5.5cm]{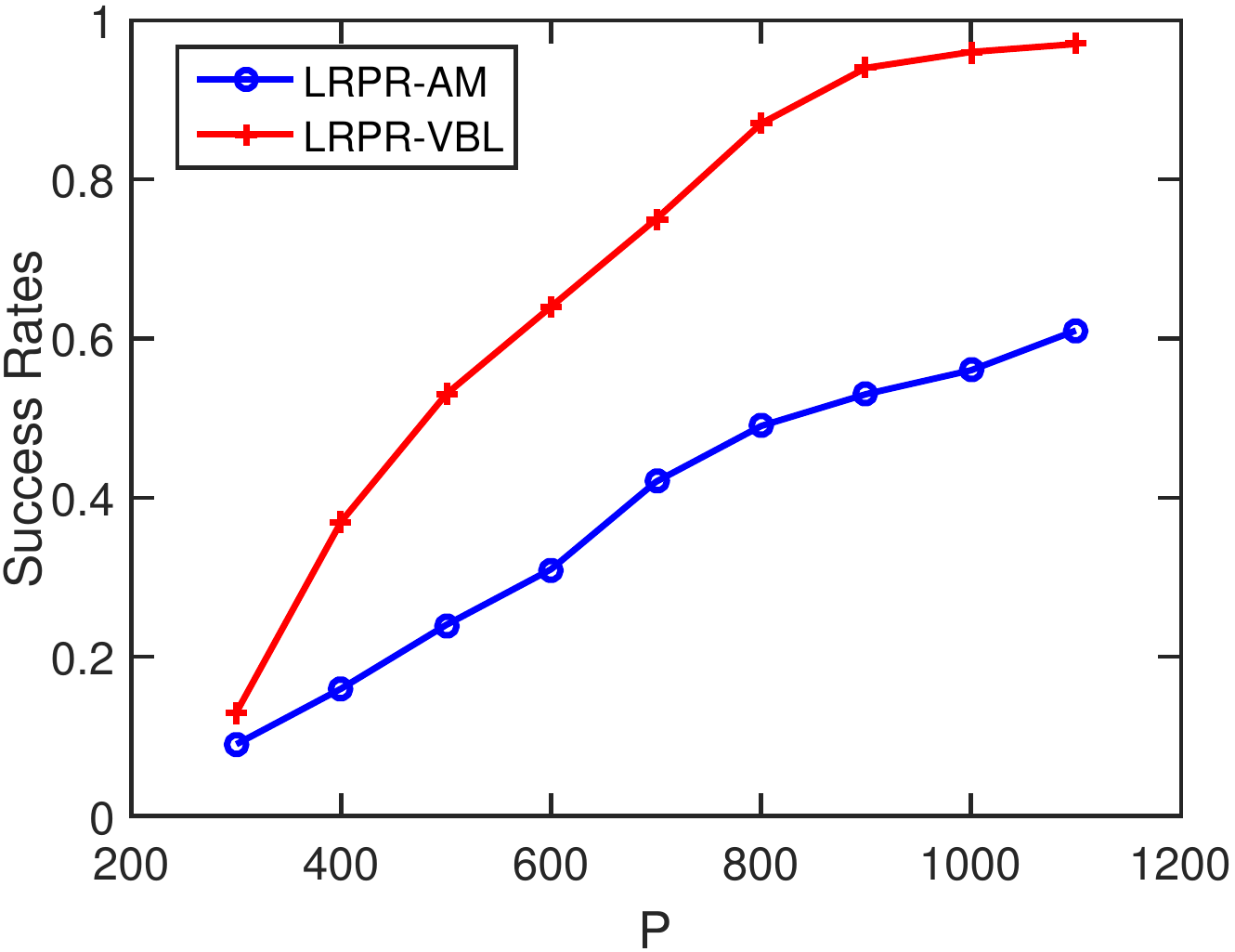}}
    \hfil
    \subfigure[$r=5$]{\includegraphics[width=5.5cm]{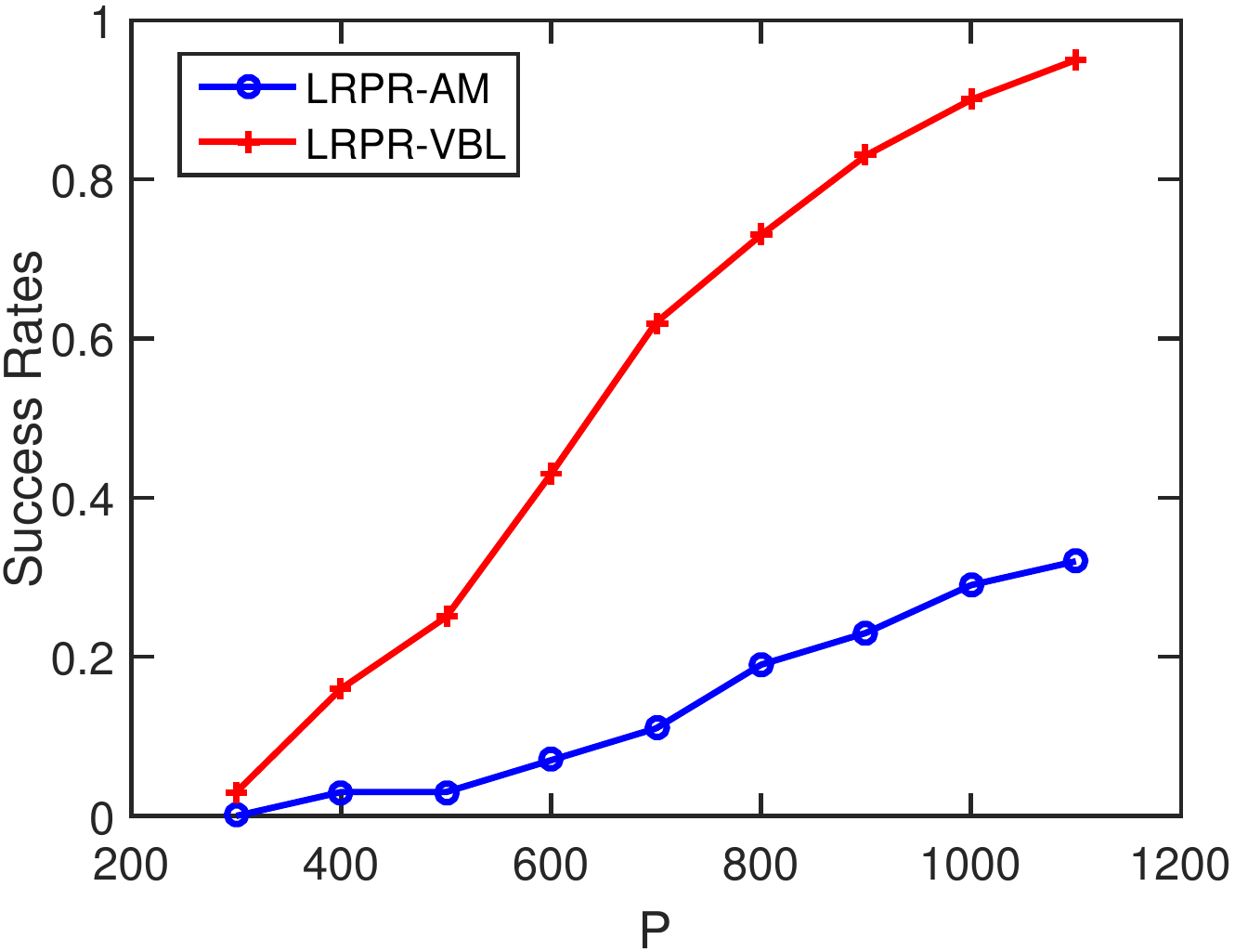}}
    \caption{Success rates of respective algorithms vs. the number of measurements $P$.}
    \label{fig2}
\end{figure*}


\subsection{Computational Complexity}
The proposed algorithm requires to calculate the inverse of a
matrix, i.e. (\ref{eq_cov}), at each iteration. Its computational
complexity scales cubically in the dimensional of the matrix,
which limits its application to large-scale problems. To address
this issue, a conjugate gradient method
\cite{Fornasier2016Conjugate} can be used to significantly
speed-up the numerical solution of the $n\times n$ linear system,
which can be straightforwardly applied to solve (\ref{eq_cov}).
Besides, a recent work studies the low complexity sparse Bayesian
learning algorithm \cite{al2018gamp}, where the expectation step
is replaced by an approximate message passing algorithm
\cite{donoho2009message,rangan2011generalized} to reduce the
computation complexity. In \cite{YangFang18}, an inverse-free
Bayesian algorithm was developed via relaxed evidence lower bound
maximization. The idea can also be applied to our problem to
circumvent the cumbersome matrix inversion operations.

\section{Simulation Results}
In this section, simulation results are provided to illustrate the
effectiveness of our proposed method (referred to as LRPR-VBL).
The parameters used in our method are set to $a = b = v =
10^{-10}$, $\boldsymbol{W} = 10^{10}\boldsymbol{I}$, as suggested
in \cite{YangFang18}. We compare our method with the
state-of-the-art low-rank phase retrieval algorithm (referred to
as LRPR-AM) \cite{VaswaniNayer17} which performs the low-rank
phase retrieval via an alternating minimization strategy. To
provide a fair comparison, both methods employ a same spectral
initialization scheme which was introduced in
\cite{VaswaniNayer17}.

We first examine the performance of respective algorithms under
different choices of rank $r$. We randomly generate a rank-$r$
matrix $\boldsymbol{X}$ of size $100\times 100$ by multiplying two
matrices $\boldsymbol{E}\in\mathbb{C}^{100\times r}$ and
$\boldsymbol{F}\in\mathbb{C}^{r\times 100}$. The entries of
$\boldsymbol{E}$ and $\boldsymbol{F}$ are sampled from a
circularly-symmetric complex normal distribution. A set of
measurement matrices $\{\boldsymbol{A}_m\}_{m=1}^{100}$ are
produced with each of them is of size $P\times 100$. Entries of
the measurement matrices are independent and follow a
circularly-symmetric complex normal distribution. Fig. \ref{fig1}
plots the success rates vs. the rank $r$ of the low-rank matrix
$\boldsymbol{X}$, where we set $P=500$, $P=600$ and $P=700$,
respectively. Results are averaged over 100 independent trails. A
trail is considered to be successful if the relative error of
the recovery is less
than 0.1, and we defined the relative error as
\begin{equation}
\text{RE} = \frac{\sum_{m=1}^{M} \min \limits_{\phi_m}
    ||\boldsymbol{x}_m-e^{j\phi_m}\hat{\boldsymbol{x}}_m||^2}{||\boldsymbol{X}||^2_F}
\end{equation}

From Fig. \ref{fig1}, we see that our proposed method presents a
clear advantage over the LRPR-AM. This performance improvement is
probably due to the fact that the EM algorithm, compared to the
alternating minimization technique, is less prone to being trapped
in undesirable local minima.

Next, we generate a matrix whose rank is fixed to be $r$. The
success rates of respective methods as a function of the number of
measurements, $P$, are plotted in Fig. \ref{fig2}, in which $r$ is
set to $3$, $4$ and $5$, respectively. From Fig. \ref{fig2}, we
see that to attain a same success recovery rate, our proposed
method needs fewer measurements than the LRPR-AM method, which,
again, demonstrates superiority of our proposed method.

\begin{figure}[t]
    \centering
    \includegraphics [width=6.5cm]{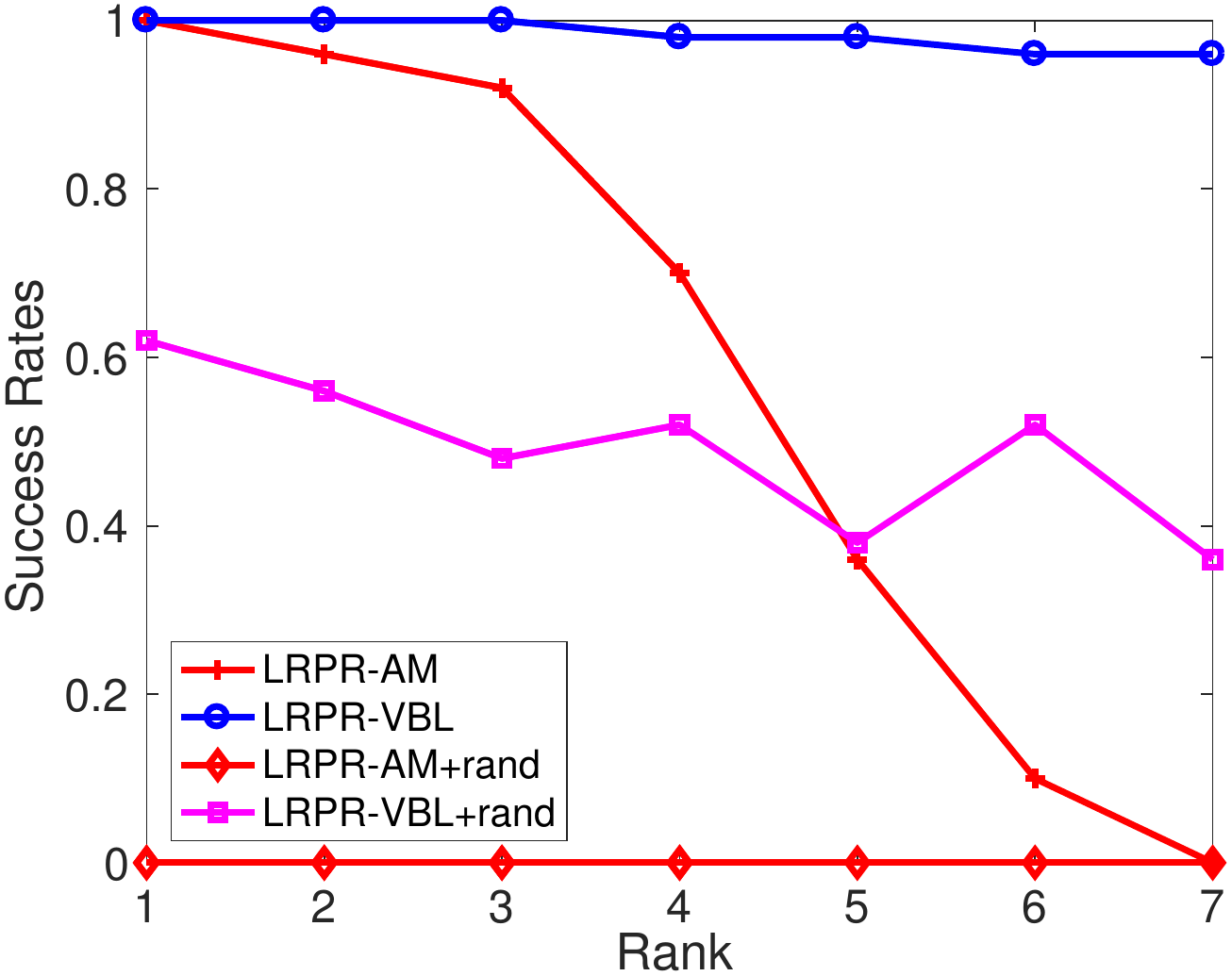}
    \caption{Success rates of respective algorithms with random
    initializations
    vs. $r$ when $P=1200$.}
    \label{fig_random}
    \end{figure}

Random initialization is a much simpler procedure and
model-agnostic, which impels its wide use in practice. This raises
an interesting question that whether our proposed LRPR-VBL
algorithm with random initialization is still effective. We
consider a setting similar to our previous experiment, and the
initialization point is chosen uniformly random. LRPR-VBL+rand
algorithm and LRPR-AM+rand stand for LRPR-VBL algorithm with
random initialization and LRPR-AM algorithm with random
initialization, respectively. We fix the dimension $P= 1200$ and
change the rank from $1$ to $7$. For each rank, we randomly
generate the sampling vector
$\boldsymbol{a}_{p,m}\sim\mathcal{CN}(\boldsymbol{0},\boldsymbol{I})$
in (\ref{lrm}). Results are averaged over $50$ independent trials.
As we can see from Fig. \ref{fig_random}, our proposed method
LRPR-VBL suffers from a certain amount of performance loss when
the initialization point is chosen randomly. But it still performs
better than the LRPR-AM method which completely fails with random
initialization points. This implies that our proposed LRPR-VBL
algorithm is less sensitive to the choice of the initialization
point.


%
%

\section{Conclusions}
The problem of low-rank phase retrieval was studied in this paper.
To estimate a complex low-rank matrix from magnitude-only
measurements, we proposed a hierarchical prior model, in which a
Gaussian-Wishart hierarchical prior is placed on the underlying
matrix to promote the low-rankness. Based on this hierarchical
prior model, we developed a variational EM method for low-rank
phase retrieval. Simulation results showed that our proposed
method offers competitive performance as compared with other
existing methods.

\bibliography{newbib}
\bibliographystyle{IEEEtran}

\end{document}